# DVIO: Depth-Aided Visual Inertial Odometry for RGBD Sensors

Abhishek Tyagi*    Yangwen Liang†    Shuangquan Wang‡    Dongwoon Bai§

SOC R&D, Samsung Semiconductor, Inc.

**ABSTRACT**

In past few years we have observed an increase in the usage of RGBD sensors in mobile devices. These sensors provide a good estimate of the depth map for the camera frame, which can be used in numerous augmented reality applications. This paper presents a new visual inertial odometry (VIO) system, which uses measurements from a RGBD sensor and an inertial measurement unit (IMU) sensor for estimating the motion state of the mobile device. The resulting system is called the depth-aided VIO (DVIO) system. In this system we add the depth measurement as part of the nonlinear optimization process. Specifically, we propose methods to use the depth measurement using one-dimensional (1D) feature parameterization as well as three-dimensional (3D) feature parameterization. In addition, we propose to utilize the depth measurement for estimating time offset between the unsynchronized IMU and the RGBD sensors. Last but not least, we propose a novel block-based marginalization approach to speed up the marginalization processes and maintain the real-time performance of the overall system. Experimental results validate that the proposed DVIO system outperforms the other state-of-the-art VIO systems in terms of trajectory accuracy as well as processing time.

## 1 INTRODUCTION

Nowadays, augmented reality (AR) is becoming a key feature for the latest mobile devices. Consequently, AR applications, such as 3D reconstruction, AR filters, in-home shopping, space exploration, AR gaming, AR meetings, etc., are shaping our post-COVID lifestyles.

In all these applications, six degrees-of-freedom (6DOF) pose tracking of the device is very critical to help localize the device in an unknown environment and render the augmentation with respect to the device. This problem has been solved using the simultaneous localization and mapping (SLAM) algorithm. Originally, the visual SLAM was used to solve this problem. Since the original monocular visual SLAM [1] work, multiple impressive implementations of the visual SLAM have shown good performance in different applications [2], [3], [4]. Unfortunately, a visual SLAM system cannot recover the absolute scale of the environment due to the use of a single monocular camera. To overcome this, a stereo camera [5] or RGBD sensor [6], [7] is introduced. Even with the scale estimation from other sources, the pose estimation in visual SLAM system gets impacted negatively when we have motion blur in images due to fast motion, or have low texture scene, which reduces the number of trackable features. These issues mainly arise from the fact that in the visual SLAM we only use two-dimensional (2D) feature measurement in the camera frame to estimate the pose. To tackle these issues, an inertial measurement unit (IMU) sensor is coupled with the camera to aid the pose estimation. Such setup is termed as Visual Inertial Odometry (VIO). In the past few years multiple monocular VIO systems have shown good performance for the pose estimation in challenging environments [8], [9]. These systems either use an Extended Kalman Filter (EKF)-based framework or a nonlinear optimization-based framework to create a tightly coupled VIO system. The main advantage of an EKF framework-based system for VIO [10], [11] is that they are able to perform good pose estimations in challenging environments with low processing requirement. However, the main drawback of the EKF-based VIO system has been its reliability. They are less robust with initialization of the filter and also have higher linearization errors in the estimated pose compared to nonlinear optimization systems.

---

*email: a.tyagi@samsung.com
†email: liang.yw@samsung.com
‡email: shuangquan.w@samsung.com
§email: dongwoon.bai@samsung.com

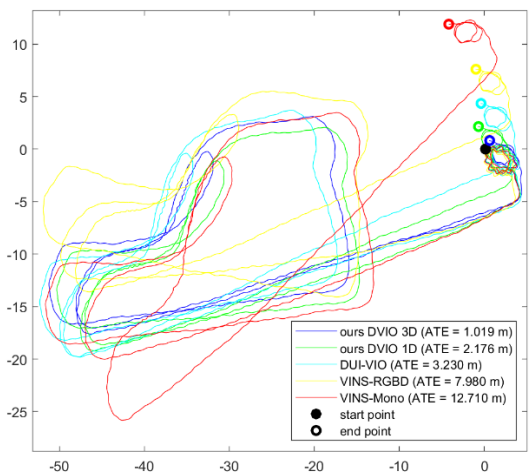

Figure 1: Trajectory comparisons of different VIO systems for the "hall2" sequence from the VCU-RVI benchmark [30]. The total duration of this sequence is 318 seconds and the total distance is 309 meters.

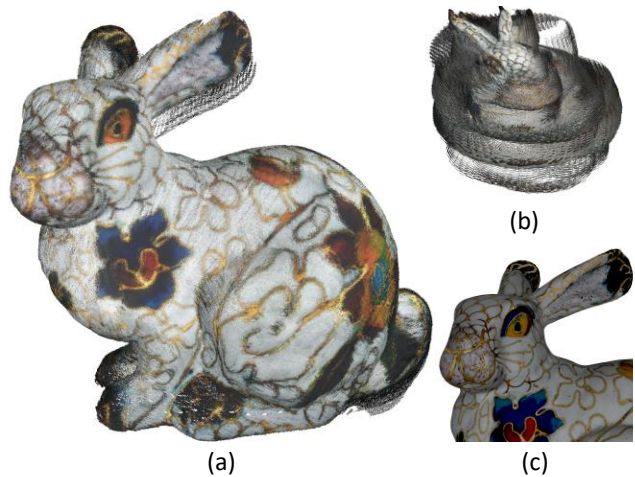

Figure 2: Stanford bunny point clouds generated from (a) the proposed DVIO 1D system and (b) the VINS-Mono system [13]. (c) shows the ground truth 3D model. VINS-RGBD [18] fails to initiate.

In the last few years the nonlinear optimization-based VIO systems [12], [13], [14] have shown good methods to perform tight coupling of the visual and inertial measurements using IMU pre-integrate [15] in a bundle adjustment [16] framework. These systems have shown better performance in the pose estimation accuracy as well as more robustness to the initialization motion needed for the system. The recent implementations [13] have also shown that by using a smart software architecture we can achieve the real-time performance on mobile devices. However, these implementations only compute the vision residue using the 2D position of the landmark in camera frame. In the case of RGBD sensors, the depth map can be used to generate extra constraints for the landmark position, and hence improve the accuracy and robustness of the optimization step. In [17], it was shown how to use an EKF framework to perform the tight coupling of the inertial, visual and depth measurements for the RGBD-IMU SLAM, and [18] has used the nonlinear optimization to perform the tight coupling of IMU and visual measurements and use depth from a RGBD sensor to improve the 3D estimation of the 2D feature. There is also recent work [19], which is using depth as a measurement in the optimization step and using feature uncertainty model to improve the pose estimation.

### 1.1 Related Work

In the early work on integrating inertial measurements with RGBD data [20], the EKF framework was used to perform a loose coupling of the pose estimated using inertial data with the pose estimated using a sparse visual-only RGBD SLAM system. In [17], the author proposed a RGBD-IMU tightly coupled EKF framework for the pose estimation as well as calibration between the two sensors. They analyzed the observability of the nonlinear system describing the RGBD-IMU calibration and 6DOF pose estimation of the IMU state. This method improved the performance of the EKF-based RGBD-IMU SLAM system and also reduced the processing time, but the EKF-based systems can have lower pose accuracy due to higher linearization error and are not robust to initialization of the state.

In [12], a keyframe-based sliding window approach for VIO was proposed using nonlinear optimization, this work was then further extended in [13], which not only improved the accuracy of the system but also improved the performance for initialization and processing time of the overall system [8]. In [18], RGBD data is used with IMU data in a tightly coupled nonlinear optimization approach. In this solution, the depth measurement of the 3D landmark is used to perform the triangulation of the feature depth in the first keyframe, in which the feature is observed. This helps in better initialization of the feature in the sliding window. In this work they also reduced the size of the nonlinear optimization problem by making the feature state for the landmark as constant for the features, which are initialized using the depth measurement. This approach resulted in more accurate and robust optimization solution for the overall sliding window. In this system the depth measurement is not used as a constraint in the optimization step. In [19], the author has proposed a system that models uncertainties in the depth measurement from a RGBD sensor using a Gaussian mixture model (GMM) and then incorporates the measurement in the initialization and optimization process of the VIO system. However, the depth measurement can be used to provide much stronger constraints for the pose estimation process that will give better pose accuracy from the nonlinear optimization step. Moreover, for a commercially realistic VIO system using RGBD sensors, any increase in the optimization problem size needs to be compensated with the improvement in the processing time.

### 1.2 Contribution

Unlike the previous proposed VIO systems, we demonstrate that depth measurements from the RGBD sensor can be utilized to provide constraints for the nonlinear optimization process with different feature parameterizations as well as unsynchronized sensors. In this work, we introduce a method to add the depth measurement in a nonlinear optimization-based VIO system. In our system, we first show how to use depth measurements in the optimization step of the VIO, and then describe a method of adding the depth measurement constraint in a sliding window that has 1D parametrized feature states. In addition, we also show the addition of the depth in a VIO system with feature state parameterized using the full 3D observation of the landmark in the first keyframe. With the 3D parametrization, we are able to utilize all the constraints provided by depth measurements for the optimization processes. We also show how depth measurements can be used in handling the scenario, where we have time offset between RGBD data and IMU data. With these changes we have increased the overall size of the optimization problem. To compensate for this increase we present a block-by-block fast marginalization scheme that significantly reduces the overall processing time for the marginalization step and maintains the real-time performance of our VIO system. The performance improvements in both pose estimation accuracy and processing time are validated using an extensive range of test cases in our experiments.

As shown in Figure 1 and Figure 2, we demonstrate the performance of our proposed system compared with other state-of-the-art VIO systems. In Figure 1, the test sequence hall2 from the VCU-RVI benchmark [30] was obtained by walking in a looped trajectory (i.e., returning to the starting point). In Figure 2, the 3D point cloud of the object in the camera view is generated by the resulting poses obtained from our proposed system and VINS-Mono. Here, the test sequence is the u3d spiral which will be introduced in Section 3. From the result, one can see the impact of good pose estimation on the final result for the 3D reconstruction application. In the next section we will go over the algorithmic details of our proposed system.

## 2 DEPTH-AIDED VISUAL INERTIAL ODOMETRY

Similar to [13] and [18], the proposed DVIO is implemented using software architecture consisting of two main threads as shown in Figure 3. The first thread is named as the frontend thread, which handles the feature tracking and the depth map processing using the data from the RGBD sensor. In this thread, we detect 2D features [22] in camera image and track them from one frame to the next frame using the KLT [23] algorithm. We also compute the depth of the 2D feature using the depth map in the RGBD data and the 2D feature location. The second thread is named as the backend thread, which handles the full state estimation using a sliding window-based nonlinear optimization. In this thread, the feature tracking data and the IMU data are used to perform the initialization of the estimator. In the DVIO implementation, the steps followed in the initialization process are the same as those defined in [13]. After initialization is successful, the sliding window is created using the IMU pose, speed and bias from the IMU pre-integration [15] and the landmark observations in the camera frames computed in the frontend thread. This sliding window is then used to perform the state estimation using the nonlinear optimization process. Once we have the estimated state of the sliding window, it is shared with the loop closure module to perform the global pose graph optimization [13]. The state vector is also used to compute the camera pose or IMU pose at the input sensor frequency using the motion-only bundle adjustment or IMU propagation as described in [13].

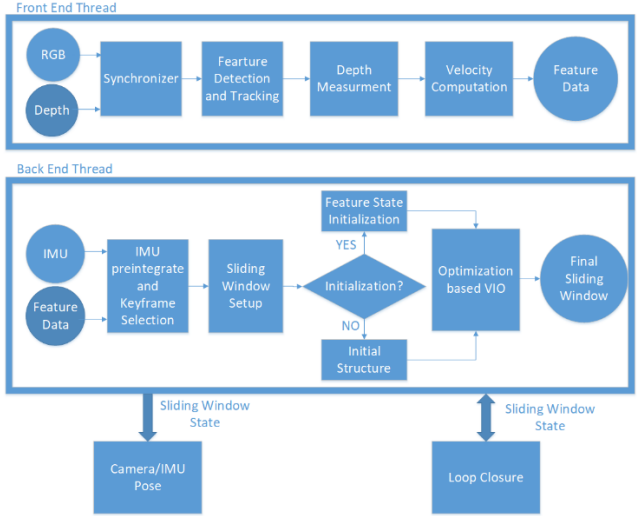

Figure 3: Block diagram illustrating the two main threads of DVIO. The front end thread processes RGBD data and creates landmark feature data, and the back end thread takes IMU data and feature data to perform optimization-based VIO and also integrate with loop closure and pose output for the sensor.

The full state vector in sliding window is defined as follows:
$$X = [x^{nav}; x^{feat}]$$
$$x^{nav} = [x_0^{nav}; x_1^{nav}; ...; x_{K-1}^{nav}] \quad (1)$$
$$x^{feat} = [x_0^{feat}; ...; x_{L-1}^{feat}]$$

where $x_k^{nav}$ is the IMU state that contains pose $x_k^R$, speed and bias $x_k^{SB}$ of the IMU at the time of $k^{th}$ keyframe, $x_l^{feat}$ is the feature state for $l^{th}$ 3D landmark, and the sliding window has $K$ keyframes and $L$ features.

Similar to [13] and [18], in the DVIO, we estimate the state vector $X$ by minimizing the Mahalanobis norm of all the measurement residuals and the residuals for the prior information from marginalization as follows.

$$\min_X \left\{ \|r_M(X_r)\|^2 + \sum_{k \in B_w} \|r_{IMU}^k\|^2_{P_{b_{k+1}}^{b_k}} + \sum_{(i,j,l) \in \mathcal{F}_w} \rho\left(\|r_{VIS}^{ijl}\|^2_Q\right) + \sum_{(i,l) \in \mathcal{F}_w} \rho\left(\|r_{VIS}^{il}\|^2_Q\right) \right\} \quad (2)$$

Here, $\rho(\cdot)$ is the Cauchy loss function, $B_w$ is the index set of sliding window, $\mathcal{F}_w$ is the set of feature observations within the sliding window, $X$ is the set of state variables to optimize over, $P_{b_{k+1}}^{b_k}$ and $Q$ are the covariance matrices used for computing Mahalanobis norm as described in [13], $r_{IMU}^k$ is the IMU residue factor, $r_{VIS}^{il}$ is the vision residue factor for the first keyframe, in which landmark is observed, $r_{VIS}^{ijl}$ is the vision residue factor for all the other keyframes, in which the landmark is observed, and $r_M(X_r)$ is the marginalization residue factor for the state $X_r$ after marginalization.

In the DVIO, the IMU residue, $r_{IMU}^k$, and the marginalization residue, $r_M$, are the same as those defined in [13] for the optimization step. For the vision residue factors, $r_{VIS}^{ijl}$ and $r_{VIS}^{il}$, we propose two solutions based on the feature parameterization in the state vector. In the 1D feature parametrization, we use the same feature state as defined in [13], [19], [24], where we parameterize a 3D landmark using the inverse depth in the first camera keyframe, in which the landmark is observed. In the 3D feature parameterization, we parameterize the 3D landmark using the inverse depth and the 2D location of the feature in the first camera keyframe, in which the landmark is observed. Since most of the current RGBD sensors used in the mobile devices are using structure light [25] technology, the noise in the depth measurement is inversely proportional to the depth of the landmark, hence using inverse depth enables us to use a Gaussian model for the depth measurement noise. In scenarios, where we have no available depth for a feature measurement due to holes in the depth map, we do not compute the depth constraint for the feature measurement.

### 2.1 1D Feature Parameterization

In the case of 1D feature parameterization, the feature state is defined using the inverse depth of the landmark in the first keyframe, in which the feature is observed. This keyframe is called as the anchor keyframe for the landmark, and all the subsequent keyframes, in which the landmarks are observed, are called as the non-anchor keyframe for the landmark. The feature state of a landmark in the sliding window is initialized by two steps. In the first step, we compute the depth of the landmark in the anchor keyframe using the landmark's depth measurements in all the keyframes and transforming them to the anchor keyframe using the keyframe pose in sliding window. The feature state is then initialized as an average of all the computed depths of the landmark in the anchor keyframe. The remaining landmark that could not be initialized in the first step are initialized using triangulation. Specifically, the Direct Linear Transformation (DLT) algorithm [26] is utilized to estimate the depth of the landmark in the anchor keyframe using the 2D feature measurement in the sliding window.

The DVIO vision residual for the non-anchor keyframe for a landmark $l$ that is observed in anchor keyframe $i$ and a non-anchor keyframe $j$, is computed as follows:

$$[\tilde{x}_{jl}, \tilde{y}_{jl}, \tilde{z}_{jl}]^T = R_{c_j c_i} \left[\frac{\bar{u}_{il}}{\lambda_l}, \frac{\bar{v}_{il}}{\lambda_l}, \frac{1}{\lambda_l}\right]^T + t_{c_j c_i}$$
$$r_{VIS}^{ijl} = \left[\frac{\tilde{x}_{jl}}{\tilde{z}_{jl}}, \frac{\tilde{y}_{jl}}{\tilde{z}_{jl}}, \frac{1}{\tilde{z}_{jl}}\right]^T - \left[\bar{u}_{jl}, \bar{v}_{jl}, \frac{1}{\bar{z}_{jl}}\right]^T \quad (3)$$

where, $\lambda_l$ is the inverse depth of the landmark $l$ in the anchor keyframe $i$, $(\bar{u}_{il}, \bar{v}_{il})$ and $(\bar{u}_{jl}, \bar{v}_{jl})$ are the measured 2D locations of the landmark in the keyframes $i$ and $j$ in the normalized homogeneous coordinate system, respectively. $R_{c_j c_i}$ and $t_{c_j c_i}$ are the rotation and translation from the keyframe $i$ to the keyframe $j$, and $\bar{z}_{jl}$ is the depth measured for the landmark $l$ in the keyframe $j$.

Next we use the depth $\bar{z}_{il}$ measured in the anchor keyframe and estimated inverse depth $\lambda_l$ in the state to generate the vision residue in the anchor keyframe $i$.

$$r_{VIS}^{il} = \lambda_l - \frac{1}{\bar{z}_{il}} \quad (4)$$

The residues $r_{VIS}^{ijl}$ and $r_{VIS}^{il}$ combine the depth measurement from the depth sensor with the feature measurement in the camera image to create the vision constraints in the DVIO.

## 2.2 3D Feature Parameterization

In the case of 3D feature parameterization, the feature state is defined by the estimate of 2D location of the feature and the inverse depth of the feature in the anchor keyframe. Hence, the feature state becomes a 3D vector, denoted as $\boldsymbol{x}_l^{feat} = [u_{il}, v_{il}, \lambda_l]^T$, where $(u_{il}, v_{il})$ is the estimated 2D location of the feature in the anchor keyframe $i$ in the normalized coordinate system. The inverse depth part of the feature state is initialized in the same way as it was done in the case of 1D feature parameterization. For the initialization of the 2D feature location of the landmark in the anchor keyframe, we utilize the DLT algorithm [26]. In the DLT algorithm using the 2D feature measurement for the landmark in the sliding window and the sliding window pose, we can compute the 2D location of the feature in the anchor keyframe in the normalized coordinate system, which is used to initialize the feature state for the landmark.

Similar to the 1D feature parameterization, we can compute the non-anchor keyframe vision residue using the new feature state as follows.

$$[x_{jl}, y_{jl}, z_{jl}]^T = R_{c_j c_i} \left[\frac{u_{il}}{\lambda_l}, \frac{v_{il}}{\lambda_l}, \frac{1}{\lambda_l}\right]^T + t_{c_j c_i} \quad (5)$$

and the residue can be obtained as

$$r_{VIS}^{ijl} = \left[\frac{x_{jl}}{z_{jl}}, \frac{y_{jl}}{z_{jl}}, \frac{1}{z_{jl}}\right]^T - \left[\bar{u}_{jl}, \bar{v}_{jl}, \frac{1}{\bar{z}_{jl}}\right]^T \quad (6)$$

In the case of anchor keyframe, we have the estimated and measured value of the landmark in the full 3D. This allows us to extend the anchor keyframe vision residue to a three dimensional residual.

$$r_{VIS}^{il} = [u_{il}, v_{il}, \lambda_l]^T - \left[\bar{u}_{il}, \bar{v}_{il}, \frac{1}{\bar{z}_{il}}\right]^T \quad (7)$$

With this we are able to use all the constraints provided by the feature and depth measurement in the sliding window. These residues are then used with the IMU residual and marginalization factor to estimate more accurate pose using the nonlinear optimization process. To perform the nonlinear optimization, we employ Ceres solver [27] in the software implementations.

## 2.3 Unsynchronized Sensors

In most of the commercially available devices that have a RGBD sensor and an IMU sensor, we find that the timestamps of IMU and RGBD sensors are synchronized at the hardware level. It makes the state estimation easier as there is no need to estimate the time offset between the two sensor modalities. In scenarios where we have temporal misalignment (i.e., time offset) between the two sensor modalities, we can use depth and feature measurement to perform the time offset estimation. In our consideration, we assume that we only have the time offset between RGBD and IMU sensors and no time offset between camera image and depth map.

Let's consider a time offset $t_d$ between the IMU timestamps $t_{IMU}$ and RGBD timestamps $t_{RGBD}$. Similar to [28], we use the general case where the time offset is constant but unknown, such that $t_{IMU} = t_{RGBD} + t_d$. The landmark's 2D feature observation in two consecutive image frames can be characterized as $(u_l^{k+1}, v_l^{k+1})$ in frame $k+1$ at time $t_{k+1}$ and $(u_l^k, v_l^k)$ in image frame $k$ at time $t_k$. Combined with the depth observations $z_l^{k+1}$ and $z_l^k$ for the same landmark from the corresponding depth map, we can obtain the 3D velocity of the feature as:

$$\boldsymbol{V}_l^k = \frac{1}{(t_{k+1} - t_k)} \left( \begin{bmatrix} u_l^{k+1} \\ v_l^{k+1} \\ z_l^{k+1} \end{bmatrix} - \begin{bmatrix} u_l^k \\ v_l^k \\ z_l^k \end{bmatrix} \right) \quad (8)$$

Using the velocity and time offset we can shift the landmark observation in a keyframe $i$ to the IMU time stamps.

$$\begin{bmatrix} \bar{u}_{il}(t_d) \\ \bar{v}_{il}(t_d) \\ \bar{z}_{il}(t_d) \end{bmatrix} = \begin{bmatrix} \bar{u}_{il} \\ \bar{v}_{il} \\ \bar{z}_{il} \end{bmatrix} + t_d \boldsymbol{V}_l^i \quad (9)$$

Thus, we have moved the observation of the landmark in the keyframe to compensate for the time offset. Subsequently, we are able to obtain the vision residue for the 1D and 3D feature parameterizations with the shifted feature observation.

- 1D parameterized feature vision residue:

$$\begin{bmatrix} \tilde{x}_{jl}(t_d) \\ \tilde{y}_{jl}(t_d) \\ \tilde{z}_{jl}(t_d) \end{bmatrix} = R_{c_j c_i} \left[\frac{\bar{u}_{il}(t_d)}{\lambda_l}, \frac{\bar{v}_{il}(t_d)}{\lambda_l}, \frac{1}{\lambda_l}\right]^T + t_{c_j c_i} \quad (10)$$

$$r_{VIS}^{ijl}(t_d) = \begin{bmatrix} \tilde{x}_{jl}(t_d)/\tilde{z}_{jl}(t_d) \\ \tilde{y}_{jl}(t_d)/\tilde{z}_{jl}(t_d) \\ 1/\tilde{z}_{jl}(t_d) \end{bmatrix} - \begin{bmatrix} \bar{u}_{jl}(t_d) \\ \bar{v}_{jl}(t_d) \\ 1/\bar{z}_{jl}(t_d) \end{bmatrix} \quad (11)$$

$$r_{VIS}^{il}(t_d) = \lambda_l - \frac{1}{\bar{z}_{il}(t_d)} \quad (12)$$

- 3D parameterized feature vision residue:

$$[x_{jl}, y_{jl}, z_{jl}]^T = R_{c_j c_i} \left[\frac{u_{il}}{\lambda_l}, \frac{v_{il}}{\lambda_l}, \frac{1}{\lambda_l}\right]^T + t_{c_j c_i} \quad (13)$$

$$r_{VIS}^{ijl}(t_d) = \left[\frac{x_{jl}}{z_{jl}}, \frac{y_{jl}}{z_{jl}}, \frac{1}{z_{jl}}\right]^T - \left[\bar{u}_{jl}(t_d), \bar{v}_{jl}(t_d), \frac{1}{\bar{z}_{jl}(t_d)}\right]^T \quad (14)$$

$$r_{VIS}^{il}(t_d) = [u_{il}, v_{il}, \lambda_l]^T - \left[\bar{u}_{il}(t_d), \bar{v}_{il}(t_d), \frac{1}{\bar{z}_{il}(t_d)}\right]^T \quad (15)$$

Here, $t_d$ is the time offset between RGBD and IMU sensors, $(\bar{u}_{il}(t_d), \bar{v}_{il}(t_d))$ and $(\bar{u}_{jl}(t_d), \bar{v}_{jl}(t_d))$ are the shifted 2D observations of the landmark $l$ in the anchor frame $i$ and the non-anchor frame $j$, $\lambda_l$ is the estimated inverse depth of the landmark $l$ in the anchor frame $i$, $(u_{il}, v_{il})$ are the estimated 2D feature observation of the landmark $l$ in the anchor frame $i$, $\bar{z}_{il}(t_d)$ and $\bar{z}_{jl}(t_d)$ are the shifted depth measurements of landmark $l$ in the anchor frame $i$ and the non-anchor frame $j$, $r_{VIS}^{il}(t_d)$ and $r_{VIS}^{ijl}(t_d)$ are the vision residues for the anchor frame and non-anchor frame using the time shifted observations, respectively. In the case of vision residue for the 3D parameterized feature, we have the estimated value of the 2D feature observation in IMU time and hence does not require to be shifted using $t_d$.

In the above vision residue with the correct value of $t_d$, the constraints can be used with the IMU constraints directly, as it will match with the IMU time stamp. To estimate the optimum value of $t_d$, similar to [28], we add $t_d$ to the state vector and iteratively optimize it with other state variable in the sliding window based optimization by minimizing following cost function:

$$\min_{X_{td}} \left\{ \|r_M(X_r)\|^2 + \sum_{k \in B_w} \|r_{IMU}^k\|_{\boldsymbol{P}_{b_{k+1}}^{b_k}}^2 \right.$$
$$\left. + \sum_{(i,j,l) \in \mathcal{F}_w} \rho \left( \|r_{VIS}^{ijl}(t_d)\|_Q^2 \right) + \sum_{(i,l) \in \mathcal{F}_w} \rho \left( \|r_{VIS}^{il}(t_d)\|_Q^2 \right) \right\} \quad (16)$$

Here, the cost function is similar to the one defined in (2) except that the vision residue $r_{VIS}^{ijl}(t_d)$ and $r_{VIS}^{il}(t_d)$ are obtained using the time shifted landmark observations and $t_d$ is added to the state vector $\boldsymbol{X}_{td}$.

## 2.4 Fast Marginalization

In the sliding window-based nonlinear optimization process, every time we have a sliding window with the max number of nodes, we need to drop certain nodes from the window to maintain the size of the problem in the optimization step. As described in [13], when we want to drop certain nodes $X_m$ and only keep remaining subset $X_r$ ( $X_m \cup X_r = X, X_m \cap X_r = \emptyset$ ), the marginalization is performed to drop the nodes $X_m$ in the optimization problem based on Schur complement [29]. Similar to [13], in DVIO for a given information matrix $H = \begin{bmatrix} H_{mm} & H_{mr} \\ H_{rm} & H_{rr} \end{bmatrix}$ and the corresponding residual vector $\vec{b} = \begin{bmatrix} \vec{b}_m \\ \vec{b}_r \end{bmatrix}$, the marginalization process computes the new information matrix and residual vector for the remaining state as follows:

$$H_{new} = H_{rr} - H_{rm}H_{mm}^{-1}H_{mr} \quad (17)$$
$$\vec{b}_{new} = \vec{b}_r - H_{rm}H_{mm}^{-1}\vec{b}_m$$

The marginalization process described in [13] uses the eigen decomposition to compute the inverse of the full matrix block $H_{mm}$, which can become very time consuming for a big matrix. For marginalization in DVIO, we utilize the sparse structure of the information matrix block $H_{mm}$. Consequently, we rearrange the elements of the matrix $H_{mm}$. First we arrange together all the sparse diagonal blocks corresponding to the landmarks that are getting marginalized. Next we arrange together the blocks corresponding to the keyframe pose $x_k^R$ and IMU constraints $x_k^{SB}$. With this the diagonal of the matrix $H_{mm}$ is now separated into two blocks: the sparse block for landmarks and the dense block for pose and IMU constraints. We handle the marginalization of these two blocks separately.

- In the case of 1D feature parameterization, the information matrix for landmarks is a diagonal matrix. Hence, its inverse can be computed easily for a full rank matrix. For a rank-deficient matrix, we use its pseudo-inverse. This will allow us to marginalize all the landmarks in one simple step.
- In the case of 3D feature parameterization, each landmark has three correlated coordinates. Hence, it is not possible to marginalize all the landmark together in one single step. Instead we compute the inverse of a $3 \times 3$ matrix and marginalize each landmark one by one. However, compared to the original method, the lower processing time can still be achieved by the well-known closed-form solution for the inverse of any $3 \times 3$ matrix with full rank. We use the pseudo-inverse to handle the rank-deficient information matrix block, arising from landmarks at infinity or landmarks visible in only one keyframe.

Once all the landmarks are marginalized, the dense block can be marginalized using the standard marginalization step. Since the size of the remaining block for marginalization is small and fixed, Schur complement can be efficiently applied using LDL decomposition [16].

With the block-based marginalization, we are able to achieve speed-up in the processing time for both 1D and 3D feature parameterized DVIOs. As a result, we are able to maintain the real-time performance of the DVIO as shown in Section 3.2. The fast marginalization algorithm is summarized in Table 1.

## 3 EXPERIMENTAL RESULTS

Our main use case for the proposed DVIO is in the augmented reality application on mobile devices. In this section, we carried out experiments to compare the performance of our proposed DVIO system with the state-of-the-art VIO systems such as VINS-Mono [13], VINS-RGBD [18] and DUI-VIO [19] on the handheld indoor test scenarios. In [8], VINS-Mono system had shown the best pose accuracy performance among all the VIO systems. Hence, we use VINS-Mono to create a baseline performance for the current monocular camera-based VIO approaches. The VINS-RGBD system added depth in the initialization step of VINS-Mono and also used depth to make the feature state constant for certain features in the optimization step. The DUI-VIO system exploited the uncertainty of the depth data and added constraints computed using the depth measurement in the optimization step. These two systems use depth from a RGBD sensor with IMU measurements to improve the pose estimation, this makes them good candidates to compare against the DVIO's performance. To evaluate the performance, we created a test case of 35 data sequences. All the data sequences are captured in a rosbag format, to enable their use with the ROS framework.

Table 1. Fast Marginalization Algorithm Summary

| | |
|---|---|
| **Input:** | Information matrix: $H$, Residual: $\vec{b}$ <br> Remaining state vector: $X_r$ <br> State vector to be marginalized: <br> $X_m = \{x_0^{feat}, ..., x_{L-1}^{feat}, x_0^{nav}\}$ |
| **Output:** | Remaining state vector: $X_r$ <br> $H_{new}$: Information matrix after marginalization. <br> $\vec{b}_{new}$: Residual after marginalization. |
| Step1: Marginalize Landmark | |
| 1 | $H = \begin{bmatrix} H_{LL} & H_{Lr_1} \\ H_{r_1 L} & H_{r_1 r_1} \end{bmatrix}, \vec{b} = \begin{bmatrix} \vec{b}_L \\ \vec{b}_{r_1} \end{bmatrix}$ |
| 2 | if (size($x_l^{feat}$) == 1) |
| 3 | $\quad H_{LL} = diag(\vec{h}_{LL})$ |
| 4 | $\quad H_{LL}^{-1} = diag(\frac{1}{\vec{h}_{LL}})$ |
| 5 | end |
| 6 | if (size($x_l^{feat}$) == 3): |
| 7 | $\quad H_{LL} = blockdiag([H_{3\times3}^0, ..., H_{3\times3}^{L-1}])$ |
| 8 | $\quad H_{LL}^{-1} = blockdiag([(H_{3\times3}^0)^{-1}, ..., (H_{3\times3}^{L-1})^{-1}])$ |
| 9 | end |
| 10 | $H_{new}^{r_1} = H_{r_1 r_1} - H_{r_1 L} H_{LL}^{-1} H_{Lr_1}$ |
| 11 | $\vec{b}_{new}^{r_1} = \vec{b}_{r_1} - H_{r_1 L} H_{LL}^{-1} \vec{b}_L$ |
| Step2: Marginalize pose, speed and bias. | |
| 1 | $H_{new}^{r_1} = \begin{bmatrix} H_{11} & H_{12} \\ H_{21} & H_{22} \end{bmatrix}, \vec{b}_{new}^{r_1} = \begin{bmatrix} \vec{b}_1 \\ \vec{b}_2 \end{bmatrix}$ |
| 2 | $H_{new} = H_{22} - H_{21} H_{11}^{-1} H_{12}$ |
| 3 | $\vec{b}_{new} = \vec{b}_2 - H_{21} H_{11}^{-1} \vec{b}_1$ |

The data sequences can be divided into three types:
- Synthetic data: The first set are the sequences created in a simulation environment using the Unity3D software. In these sequences, the camera moves in a photorealistic synthetic environment created using 3D models rendered in Unity3D. The device poses, trajectories and IMU data are obtained with similar technique introduced in [31]. The RGBD data are generated noise free. These sequences will be available for the community for future researches.
- VINS-RGBD dataset: The second set of sequences used for evaluation were created by the VINS-RGBD [18] project. These sequences were captured by handheld using an Intel RealSense D435i sensor inside a large room with three motion scenarios.
- VCU-RVI handheld benchmark: The third set of sequences were created by the VCU-RVI benchmark [30]. In this set of sequences, the data is captured using an Occipital Structure Core sensor with different motion scenarios. There are total twenty-seven data sequences covering a total of ~2.7 kilometers trajectory were recorded in various indoor environments.

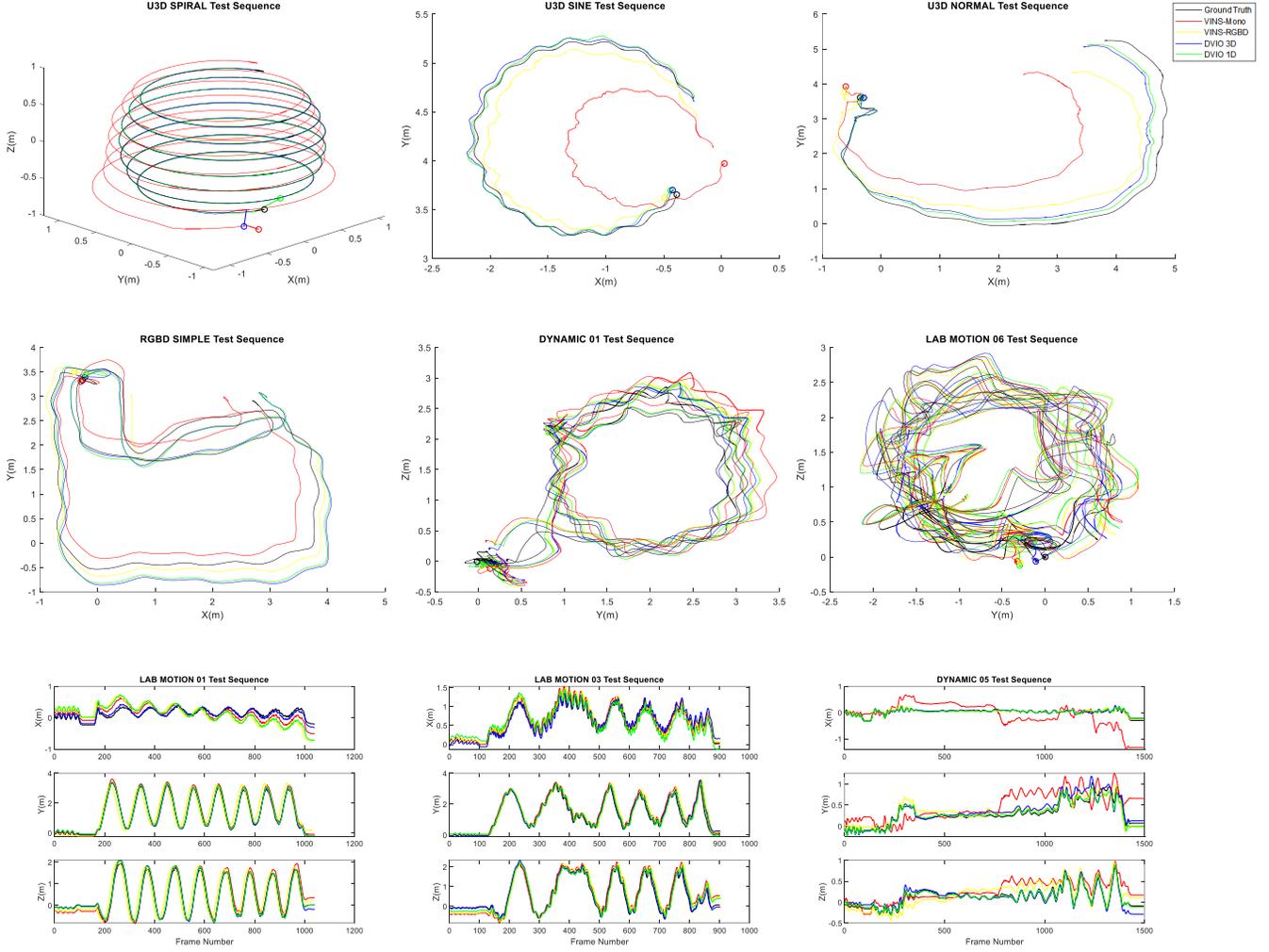

Figure 4: Trajectory Comparison for test sequences. First row is for simulation sequences, second and third row is for VINS-RGBD and DUI-VIO sequences. The circles on the 3D trajectories in the first two rows shows the starting point for the motion sequence.

In the test case, we have 6 sequences (rgbd simple, rgbd normal, rgbd rotation, u3d simple, u3d normal, u3d rotation) with unsynchronized time stamps between RGBD and IMU sensors. With these sequences we are able compare all the aspects of DVIO with other state-of-the-art VIO systems.

### 3.1 Performance Evaluation

To compare the pose accuracy of different systems on the test sequences, we use RMSE of the absolute trajectory error (ATE) and relative pose error (RPE) [32], [33] for the estimated trajectory with respect to the ground truth trajectory.

- ATE:

$$RMSE_{ATE} = \sqrt{\frac{1}{N}\sum_{n=0}^{N-1}\|trans(Q_n^{-1}\,S\,P_n)\|^2} \quad (18)$$

- RPE for every other frame using consecutive frame pairs:

$$E_n = (Q_n^{-1}\,Q_{n+1})^{-1}(P_n^{-1}P_{n+1}) \quad (19)$$

$$RMSE_{RPE} = \sqrt{\frac{1}{N-1}\sum_{n=0}^{N-2}\|trans(E_n)\|^2} \quad (20)$$

Here, $Q_n$ is the ground truth pose, $P_n$ is the estimated pose for the frame $n$ in the sequence, and $S$ is the rigid body transformation between the ground truth and estimated trajectory. $N$ is the total number of frames in the test sequence and $trans(E_n)$ refers to the translational components of the pose error $E_n$.

The result of the tests are summarized in Table 2 and Table 3. The smallest RMSE is highlighted in boldface for every sequence. In the table an "X" indicates that the method failed in initialization or resulted in a large (>50 meters) ATE. In Table 2, we present the results for the synthetic dataset and the VINS-RGBD dataset. For all the synthetic data sequences the performance of DVIO is better than the other systems. This shows that in scenario with no measurement noise DVIO system will have best pose accuracy. In VINS-RGBD dataset the user is moving around in a very large room, pointing the camera towards objects that are far away and performing quick motions. Due to these factors we observe a higher error in depth measurement for these sequences. In VINS-RGBD, the depth measurement is only used to initialize the feature state, hence its final estimated pose is less impacted by higher depth measurement errors. In contrast, we use the depth measurement to compute constraints in the optimization step. Due to this reason, we observe higher errors for DVIO systems in the case of "rgbd" sequences as compared to VINS-RGBD, but the performance gap is small.

Table 2. RMSE results for ATE and RPE (in meter) for the data sequences. Unity 3D sequences' names start with u3d, VINS-RGBD sequences' names start with rgbd.

| Dataset | ATE | | | | RPE | | | |
|---|---|---|---|---|---|---|---|---|
| | VINS-Mono | VINS-RGBD | DVIO 1D | DVIO 3D | VINS-Mono | VINS-RGBD | DVIO 1D | DVIO 3D |
| u3d spiral | 0.145 | X | **0.032** | 0.046 | 0.1502 | X | 0.1445 | **0.1444** |
| u3d sine | 0.353 | 0.084 | **0.036** | 0.037 | **0.0240** | 0.0259 | 0.0268 | 0.0268 |
| u3d simple | 0.575 | 0.369 | 0.241 | **0.213** | **0.0541** | 0.0563 | 0.0568 | 0.0567 |
| u3d normal | 0.795 | 0.453 | **0.122** | 0.189 | **0.0585** | 0.0590 | 0.0607 | 0.0605 |
| u3d rotation | 0.924 | 0.682 | 0.836 | **0.496** | 0.0493 | 0.0510 | 0.0477 | **0.0477** |
| rgbd simple | 0.228 | **0.183** | 0.184 | 0.205 | **0.0534** | 0.0598 | 0.0594 | 0.0593 |
| rgbd normal | 0.229 | **0.059** | 0.181 | 0.204 | **0.0579** | 0.0608 | 0.0625 | 0.0628 |
| rgbd rotation | 0.196 | **0.126** | 0.195 | 0.210 | **0.0472** | 0.0481 | 0.0494 | 0.0496 |
| Overall Mean | 0.4714 | 0.2794 | 0.2564 | **0.2220** | **0.0492** | 0.0516 | 0.0519 | 0.0519 |

In Table 3, we present the results for VCU-RVI handheld benchmark dataset. In these results, easy and motion sequences have relatively simple motion scenarios and DVIO has performed with the best ATE for these sequences. Corridor and hall sequences have motion scenarios with longer trajectories, which is causing higher ATE for all the systems. In the corridor1 sequence, both DVIO 1D and DVIO 3D diverge at some point in the sequence, and in the hall3 sequence, VINS-Mono, VINS-RGBD and DVIO-1D are diverging. Overall for these longer trajectory sequences, DVIO is able to perform best in 5 out 7 sequences. In case of light and dynamic test sequences we have challenging motion scenarios with moving objects in scene and also change in lighting conditions. DVIO performs the best for all the sequences except for the light6 sequence. In case of DVIO we use depth measurement for providing constraints to the optimization step, this improves the performance of the system in scenarios with lack of sufficient visual constraints. From the results we can see that the DVIO 1D and DVIO 3D systems perform better than other systems in 29 out of 35 sequences for ATE and 23 out of 35 sequence for RPE. Also to compare the reliability of the systems, we can see that DVIO 3D only failed in one sequence and DVIO 1D had two failures, compared to 4 failures in DUI-VIO, 4 failures in VINS-RGBD and 3 failures in VINS-Mono. It confirms that the DVIO generated more accurate and reliable pose estimations compared to the other state-of-the-art systems. In Figure 4, we show the trajectory comparison of the VIO system for different test sequences.

Table 3. RMSE results for ATE and RPE (in meter) for the VCU-RVI handheld benchmark dataset

| Dataset | ATE | | | | | RPE | | | | |
|---|---|---|---|---|---|---|---|---|---|---|
| | DUI VIO | VINS-Mono | VINS-RGBD | DVIO 1D | DVIO 3D | DUI VIO | VINS-Mono | VINS-RGBD | DVIO 1D | DVIO 3D |
| corridor1 | **3.230** | 4.390 | 5.130 | X | X | 0.16 | **0.12** | 0.20 | X | X |
| corridor2 | 3.230 | 1.610 | 1.810 | 1.410 | **1.140** | 0.14 | **0.07** | 0.13 | 0.14 | 0.11 |
| corridor3 | 5.230 | 3.970 | 6.810 | 2.350 | **1.990** | 0.22 | 0.18 | 0.27 | 0.22 | **0.17** |
| corridor4 | 2.200 | 4.330 | 1.950 | 1.510 | **0.915** | 0.09 | 0.18 | 0.14 | 0.14 | **0.08** |
| hall1 | 1.510 | 4.800 | 2.340 | **0.694** | 2.360 | 0.11 | 0.17 | 0.21 | **0.06** | 0.10 |
| hall2 | 3.230 | 12.710 | 7.980 | 2.176 | **1.019** | 0.13 | 0.55 | 0.33 | 0.18 | **0.09** |
| hall3 | **6.520** | X | X | X | 6.610 | **0.35** | X | X | X | 0.69 |
| Mean | 3.0800 | 5.4840 | 4.1780 | 1.6280 | **1.4848** | 0.1380 | 0.2300 | 0.2160 | 0.1480 | **0.1100** |
| dynamic1 | 0.235 | 0.287 | 0.216 | **0.157** | 0.159 | 0.0091 | 0.0094 | 0.0091 | 0.0090 | **0.0089** |
| dynamic2 | 0.215 | 0.550 | 0.324 | **0.206** | 0.303 | **0.0062** | 0.0113 | 0.0077 | 0.0068 | 0.0098 |
| dynamic3 | 0.471 | 0.653 | 0.520 | 0.294 | **0.293** | 0.0068 | 0.0097 | 0.0068 | 0.0075 | **0.0056** |
| dynamic4 | 3.600 | 1.700 | 2.500 | **0.820** | 1.186 | 0.0193 | 0.0275 | 0.0171 | **0.0159** | 0.0169 |
| dynamic5 | 0.073 | 0.469 | 0.177 | **0.060** | 0.118 | 0.0081 | 0.0129 | 0.0089 | **0.0081** | 0.0091 |
| light1 | X | X | X | 0.871 | **0.361** | X | X | X | 0.0264 | **0.0157** |
| light2 | X | 0.400 | 0.480 | **0.370** | 0.375 | X | 0.0848 | 0.0822 | **0.0721** | 0.0750 |
| light3 | 0.378 | 0.384 | X | **0.139** | 0.156 | 0.0221 | 0.0233 | X | **0.0219** | 0.0223 |
| light4 | 0.270 | 0.290 | 3.110 | **0.175** | 0.181 | 0.0068 | 0.0073 | 0.0072 | 0.0072 | **0.0066** |
| light5 | X | X | 1.120 | 1.211 | **0.574** | X | X | 0.0290 | 0.0165 | **0.0153** |
| light6 | X | **1.060** | X | 2.200 | 1.980 | X | 0.0302 | X | **0.0214** | 0.0301 |
| Mean | 0.8107 | 0.6582 | 1.1412 | **0.2853** | 0.3733 | 0.0094 | 0.0130 | 0.0095 | **0.0091** | 0.0095 |
| easy1 | 0.109 | 0.142 | 0.137 | **0.075** | 0.079 | 0.0053 | 0.0054 | 0.0057 | 0.0048 | **0.0046** |
| easy2 | 0.151 | 0.175 | 0.431 | **0.067** | 0.073 | 0.0056 | 0.0059 | 0.0069 | 0.0057 | **0.0053** |
| easy3 | 0.123 | 0.236 | 0.116 | **0.079** | 0.084 | 0.0134 | 0.0135 | 0.0136 | 0.0133 | **0.0132** |
| motion1 | 0.532 | 0.526 | 0.581 | **0.292** | 0.371 | **0.0187** | 0.0196 | 0.0188 | 0.0187 | 0.0187 |
| motion2 | 0.565 | 0.699 | 0.660 | **0.333** | 0.475 | 0.0396 | 0.0396 | **0.0395** | 0.0400 | 0.0405 |
| motion3 | 0.253 | 0.452 | 0.361 | **0.158** | 0.161 | 0.0090 | 0.0090 | 0.0091 | **0.0083** | 0.0085 |
| motion4 | 0.550 | 0.540 | 0.488 | **0.334** | 0.345 | 0.0101 | 0.0102 | 0.0105 | 0.0101 | **0.0098** |
| motion5 | 0.372 | 0.373 | 0.378 | 0.250 | **0.226** | 0.0134 | 0.0135 | 0.0137 | **0.0133** | 0.0136 |
| motion6 | 0.592 | 0.707 | 0.725 | 0.246 | **0.225** | 0.0111 | 0.0114 | 0.0114 | **0.0110** | 0.0110 |
| Mean | 0.3608 | 0.4278 | 0.4308 | **0.2038** | 0.2266 | 0.0140 | 0.0142 | 0.0144 | 0.0139 | **0.0139** |
| Overall Mean | 1.1756 | 1.7610 | 1.5807 | **0.5843** | 0.5852 | 0.0436 | 0.0678 | 0.0633 | 0.0460 | **0.0366** |

In the evaluation experiments we have observed that the main weakness in the DVIO system comes from the depth measurement error in the RGBD sensor. The accuracy of the depth measurement in a RGBD sensor is good for depth range of 0.5 to 4 meters (room-scale), but the accuracy of the depth measurement deteriorates considerably as the scene depth increases outside of room-scale range. We also notice that in the scenario with a lot of occlusions in the scene, the depth map will have shadow regions with no depth measurement or very erroneous depth measurement. Since in the proposed approach we use depth measurements to compute constraints in the optimization step, such big measurement errors will have bigger negative impact on the DVIO performance. Hence, these scenarios need to be handled to further improve the performance of a depth-aided VIO system. This can be done by improving the depth measurement accuracy at the RGBD sensor level or by using deep learning methods.

In these experiments we also observe that the quantitative difference is small between the performance of DVIO 1D and DVIO 3D. However, we do observe that DVIO 3D has the most reliable performance with only one failure case and also consistently performs the best in longer trajectory sequences. In DVIO 3D, we have bigger state size and also compute more constraints in the optimization step as compared to DVIO 1D. These extra constraints help in the better reliability and accuracy performance in the case of DVIO 3D, however this also causes an increase in the processing time. Hence, we recommend using the DVIO 1D system in the small-scale SLAM scenarios with stringent processing time requirements, whereas in the large-scale indoor SLAM scenarios, we recommend using the DVIO 3D system for more robust and accurate performance.

### 3.2 Runtime Evaluation

In the DVIO by adding the depth measurement as part of the optimization step and also increasing the state size for the 3D case, we have increased the problem size for the nonlinear optimization process. To compensate for this, we have improved the processing time for the marginalization step in DVIO. In Table 4, we compare the total processing time and the marginalization time in a sliding window optimization for DVIO 1D with and without the fast marginalization, DVIO 3D with and without the fast marginalization, VINS-Mono and VINS-RGBD.

Table 4. Runtime Comparison (ms)

|  | Total Time | Marginalization Time |
|---|---|---|
| DVIO 1D (With Fast Marginalization) | 54.09 | 4.93 |
| DVIO 3D (With Fast Marginalization) | 83.52 | 25.46 |
| DVIO 1D (Without Fast Marginalization) | 76.60 | 27.54 |
| DVIO 3D (Without Fast Marginalization) | 464.92 | 407.33 |
| VINS-Mono | 62.91 | 26.64 |
| VINS-RGBD | 61.40 | 26.71 |

These are average runtime numbers, generated using the offline processing of the u3d sine test sequence on a Linux machine running on an Intel Core i7-5930K processor with 64 GB DRAM. As one can see from the results, the fast marginalization provides 80% and 94% improvements for the marginalization step in the 1D and 3D parametrized feature scenarios, respectively. This enables the DVIO optimization thread to run at the 10 Hz update, allowing for a real-time performance of the backend thread.

### 4 CONCLUSION

In this paper, we proposed a method to add the depth measurement from RGBD sensor in a nonlinear optimization-based VIO system. We have shown that the depth measurement can be used in both 1D and 3D feature parameterizations. Moreover, we proposed a method to utilize the depth measurement in the estimation of time offset for the unsynchronized RGBD and IMU sensors. In the marginalization process, we have also proposed a block-based fast marginalization scheme, which reduces the runtime significantly. Finally, we carried out extensive experiments to show that our proposed DVIO system outperforms the state-of-the-art in terms of reliability, accuracy and runtime.

### REFERENCES

[1] A. J. Davison, I. D. Reid, N. D. Molton and O. Stasse, "MonoSLAM: Real-Time Single Camera SLAM," in IEEE Transactions on Pattern Analysis and Machine Intelligence, vol. 29, no. 6, pp. 1052-1067, June 2007.

[2] Engel J., Schöps T., Cremers D. (2014) LSD-SLAM: Large-Scale Direct Monocular SLAM. In: Fleet D., Pajdla T., Schiele B., Tuytelaars T. (eds) Computer Vision – ECCV 2014. ECCV 2014.

[3] R. Mur-Artal, J. M. M. Montiel and J. D. Tardós, "ORB-SLAM: A Versatile and Accurate Monocular SLAM System," in IEEE Transactions on Robotics, vol. 31, no. 5, pp. 1147-1163, Oct. 2015.

[4] Duo J., Zhao L., Mao J. (2019) Sliding Window Based Monocular SLAM Using Nonlinear Optimization. In: Jia Y., Du J., Zhang W. (eds) Proceedings of 2018 Chinese Intelligent Systems Conference. Lecture Notes in Electrical Engineering, vol 529. Springer, Singapore.

[5] Lemaire, T., Berger, C., Jung, IK. et al. Vision-Based SLAM: Stereo and Monocular Approaches. Int J Comput Vision 74, 343–364 (2007).

[6] F. Steinbrücker, J. Sturm and D. Cremers, "Real-time visual odometry from dense RGB-D images," 2011 IEEE International Conference on Computer Vision Workshops (ICCV Workshops), 2011.

[7] F. Endres, J. Hess, J. Sturm, D. Cremers and W. Burgard, "3-D Mapping With an RGB-D Camera," in IEEE Transactions on Robotics, vol. 30, no. 1, pp. 177-187, Feb. 2014.

[8] J. Delmerico and D. Scaramuzza, "A Benchmark Comparison of Monocular Visual-Inertial Odometry Algorithms for Flying Robots," 2018 IEEE International Conference on Robotics and Automation (ICRA), 2018, pp. 2502-2509.

[9] H. Zhang, L. Jin, H. Zhang and C. Ye, "A Comparative Analysis of Visual-Inertial SLAM for Assisted Wayfinding of the Visually Impaired," 2019 IEEE Winter Conference on Applications of Computer Vision (WACV), 2019, pp. 210-217.

[10] A. I. Mourikis and S. I. Roumeliotis, "A Multi-State Constraint Kalman Filter for Vision-aided Inertial Navigation," Proceedings 2007 IEEE International Conference on Robotics and Automation, 2007, pp. 3565-3572.

[11] M. Li and A. I. Mourikis, "Vision-aided inertial navigation for resource-constrained systems," 2012 IEEE/RSJ International Conference on Intelligent Robots and Systems, 2012, pp. 1057-1063.

[12] S. Leutenegger et al., "Keyframe-Based Visual-Inertial SLAM using Nonlinear Optimization", Proc. of Robotis Science and Sys, 2013.

[13] T. Qin, P. Li and S. Shen, "VINS-Mono: A Robust and Versatile Monocular Visual-Inertial State Estimator," in IEEE Transactions on Robotics, vol. 34, no. 4, pp. 1004-1020, Aug. 2018.

[14] T. Qin, J. Pan, S. Cao, and S. Shen, "A general optimization-based framework for local odometry estimation with multiple sensors," 2019, arXiv:1901.03638

[15] C. Forster, L. Carlone, F. Dellaert and D. Scaramuzza, "On-Manifold Preintegration for Real-Time Visual-Inertial Odometry," in IEEE Transactions on Robotics, vol. 33, no. 1, pp. 1-21, Feb. 2017.

[16] B. Triggs, P. F. McLauchlan, R. I. Hartley and A. W. Fitzgibbon, "Bundle adjustment: A modern synthesis", Proc. Int. Workshop Vis. Algorithms, pp. 298-372, 1999.

[17] C. X. Guo and S. I. Roumeliotis, "IMU-RGBD camera 3D pose estimation and extrinsic calibration: Observability analysis and consistency improvement," 2013 IEEE International Conference on Robotics and Automation, 2013, pp. 2935-2942.

[18] Z. Shan, R. Li and S. Schwertfeger, "RGBD-Inertial Trajectory Estimation and Mapping for Ground Robots", Sensors, vol. 19, no. 10, pp. 2251, 2019.

[19] H. Zhang and C. Ye, "DUI-VIO: Depth Uncertainty Incorporated Visual Inertial Odometry based on an RGB-D Camera," 2020 IEEE/RSJ International Conference on Intelligent Robots and Systems (IROS), 2020, pp. 5002-5008.

[20] N. Brunetto, S. Salti, N. Fioraio, T. Cavallari and L. D. Stefano, "Fusion of Inertial and Visual Measurements for RGB-D SLAM on Mobile Devices," 2015 IEEE International Conference on Computer Vision Workshop (ICCVW), 2015, pp. 148-156.

[21] T. Laidlow, M. Bloesch, W. Li and S. Leutenegger, "Dense RGB-D-inertial SLAM with map deformations," 2017 IEEE/RSJ International Conference on Intelligent Robots and Systems (IROS), 2017, pp. 6741-6748.

[22] Jianbo Shi and Tomasi, "Good features to track," 1994 Proceedings of IEEE Conference on Computer Vision and Pattern Recognition, 1994, pp. 593-600.

[23] B. D. Lucas, T. Kanade et al., "An iterative image registration technique with an application to stereo vision".

[24] D. Gutiérrez-Gómez, W. Mayol-Cuevas and J. J. Guerrero, "Inverse depth for accurate photometric and geometric error minimisation in RGB-D dense visual odometry," 2015 IEEE International Conference on Robotics and Automation (ICRA), 2015, pp. 83-89.

[25] D. Scharstein and R. Szeliski, "High-Accuracy Stereo Depth Maps Using Structured Light", Proc. IEEE Conf. Computer Vision and Pattern Recognition, vol. 1, pp. 195-202, 2003-June.

[26] R. Hartley and A. Zisserman, Multiple view geometry in computer vision., Cambridge university press, 2003.

[27] S. Agarwal, K. Mierle et al., "Ceres solver", 2012.

[28] T. Qin and S. Shen, "Online Temporal Calibration for Monocular Visual-Inertial Systems," 2018 IEEE/RSJ International Conference on Intelligent Robots and Systems (IROS), 2018, pp. 3662-3669.

[29] G. Sibley, L. Matthies and G. Sukhatme, "Sliding window filter with application to planetary landing", J. Field Robot., vol. 27, no. 5, pp. 587-608, Sep. 2010.

[30] H. Zhang, L. Jin and C. Ye, "The VCU-RVI Benchmark: Evaluating Visual Inertial Odometry for Indoor Navigation Applications with an RGB-D Camera," 2020 IEEE/RSJ International Conference on Intelligent Robots and Systems (IROS), 2020, pp. 6209-6214.

[31] W. Li et al., "InteriorNet: Mega-scale Multi-sensor Photo-realistic Indoor Scenes Dataset", British Machine Vision Conference (BMVC), 2018.

[32] J. Sturm, N. Engelhard, F. Endres, W. Burgard and D. Cremers, "A benchmark for the evaluation of RGB-D SLAM systems," 2012 IEEE/RSJ International Conference on Intelligent Robots and Systems, 2012, pp. 573-580.

[33] Grupp, Michael, "evo: Python package for the evaluation of odometry and SLAM." 2017. https://github.com/MichaelGrupp/evo